\documentclass[11pt,a4paper]{article}
\usepackage{acl2018}
\usepackage{times}
\usepackage{latexsym}
\usepackage{url}
\usepackage{graphicx}
\usepackage{amsmath}
\usepackage{amssymb}
\usepackage{subfigure}
\usepackage{caption}
\usepackage{algorithm}
\usepackage{algorithmicx}
\usepackage{algpseudocode}
\usepackage{multicol}
\usepackage{multirow}
\usepackage{booktabs}
\usepackage{listings}

\aclfinalcopy % Uncomment this line for the final submission
\def\aclpaperid{279} %  Enter the acl Paper ID here

\title{Denoising Distant Supervision for Relation Extraction \\ via Instance-Level Adversarial Training}

\author{Xu Han$^{1}$, Zhiyuan Liu$^{1}$, Maosong Sun$^{1,2}$\\
$^{1}$Department of Computer Science and Technology,\\
State Key Lab on Intelligent Technology and Systems,\\
National Lab for Information Science and Technology, Tsinghua University, Beijing, China\\
$^{2}$Beijing Advanced Innovation Center for Imaging Technology,\\
Capital Normal University, Beijing, China\\
}

\date{}

\begin{document}
\maketitle
\begin{abstract}
Existing neural relation extraction (NRE) models rely on distant supervision and suffer from wrong labeling problems. In this paper, we propose a novel adversarial training mechanism over instances for relation extraction to alleviate the noise issue. As compared with previous denoising methods, our proposed method can better discriminate those informative instances from noisy ones. Our method is also efficient and flexible to be applied to various NRE architectures. As shown in the experiments on a large-scale benchmark dataset in relation extraction, our denoising method can effectively filter out noisy instances and achieve significant improvements as compared with the state-of-the-art models.
\end{abstract}

\section{Introduction}

Relation extraction (RE) aims to extract relational facts from plain text via categorizing semantic relations between entities contained in text. For example, we can extract the fact (\emph{Mark Twain}, \texttt{PlaceOfBirth}, \emph{Florida}) from the sentence ``\emph{Mark Twain} was born in \emph{Florida}''. Many efforts have been devoted to RE, either early works based on handcrafted features \cite{zelenko2003kernel,mooney2006subsequence} or recent works based on neural networks \cite{zeng2014relation,santos2015classifying}. These models all follow a supervised learning approach, which is effective, but the requirement to high-quality annotated data is a major bottleneck in practice.

\begin{figure}[t]
\centering
\includegraphics[width=0.9\linewidth]{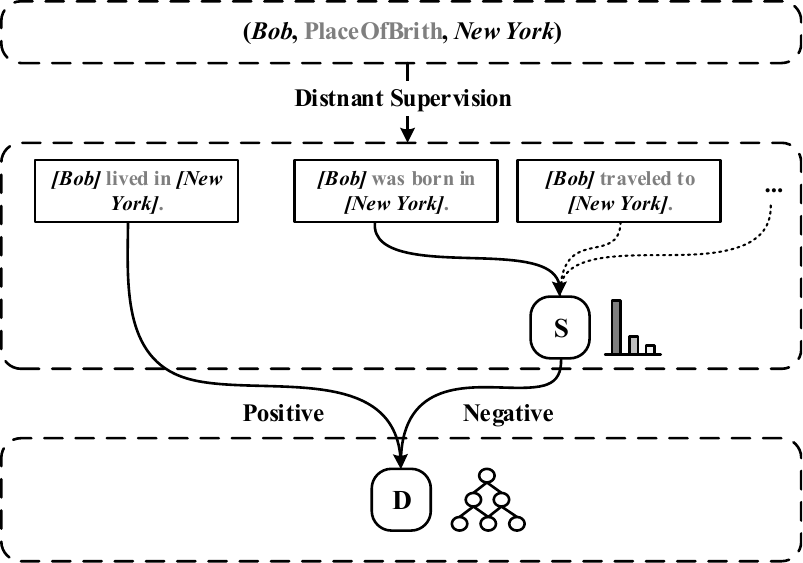}
\caption{An example of distant supervision and adversarial denoising relation extraction.}
\label{fig:examples}
\end{figure}

It is time-consuming and human-intensive to manually annotate large-scale training data. Hence, \newcite{mintz2009distant} propose \emph{distant supervision} to automatically generate training sentences via aligning KGs and text. As shown in Figure \ref{fig:examples}, distant supervision assumes that if there is a relation between two entities in a KG, all sentences that contain the two entities will be labeled with that relation. Distant supervision is an effective approach to automatically obtain training data, but it inevitably suffers from wrong labeling problems. 

To address the wrong labeling problem, \newcite{riedel2010modeling} propose multi-instance learning (MIL), and \newcite{zeng2015distant} extend the idea of MIL to neural models. \newcite{lin2016neural} further propose a neural attention scheme over multiple instances to reduce the weights of noisy instances. These methods achieve significant improvements in RE, however, still far from satisfactory. The reason is that most denoising methods simply calculate soft weights for each sentence in an unsupervised manner, which can only make a coarse-grained distinction between informative and noisy instances. Moreover, these methods cannot well cope with those entity pairs with insufficient sentences.

In order to better discriminate informative and noisy instances, inspired by the idea of adversarial learning \cite{goodfellow2014generative}, we apply adversarial training over instances to enhance RE performance. The idea of adversarial training was explored in relation extraction by generating adversarial examples with a perturbation added to sentence embeddings \cite{wu2017adversarial}, which do not necessarily correspond to real-world sentences. On contrary, we generate adversarial examples by sampling from existing training data, which may better locate real-world noise.

Our method contains two modules: a \textbf{discriminator} and a \textbf{sampler}, and the method will split the distantly supervised data into two parts, the confident part and the unconfident part. The \textbf{discriminator} is applied to judge which sentences are more likely to be annotated correctly, with the confident data as positive instances and the unconfident data as negative instances. The \textbf{sampler} module is used to select the most confusing sentences from unconfident data to cheat the discriminator as much as possible. Moreover, during several training epochs, we also dynamically select most informative and confident instances from the unconfident set to the confident set, so as to enrich the training instances for the discriminator.

The discriminator and the sampler are trained adversarially. As shown in Figure \ref{fig:examples}, during the training process, the actions of the sampler will admonish the discriminator to focus on improving those most confusing instances. Since noisy instances are ineffective to decrease the loss functions of both sampler and discriminator, the noise will be gradually filtered out during the adversarial training. Finally, the sampler can effectively distinguish those informative instances from the unconfident data, and the discriminator can well categorize relations between entities in text. As compared with the aforementioned MIL denoising methods, our method achieves more efficient noise detection in finer granularity.

We conduct experiments on a real-world dataset derived from New York Times (NYT) corpus and Freebase. Experimental results demonstrate that our adversarial denoising method effectively reduces noise and significantly outperforms other baseline methods.

\section{Related Works}

\subsection{Relation Extraction}
Relation extraction is an important task in NLP, which aims to extract relational facts from text corpora. Many efforts are devoted to RE, especially in supervised RE, such as early kernel-based models \cite{zelenko2003kernel,guodong2005exploring,mooney2006subsequence}. \newcite{mintz2009distant} align plain text with KGs and propose a distantly supervised RE model, by assuming all sentences that mention two entities can describe their relations in KGs. 

However, distant supervision inevitably accompanies with the wrong labeling problem. \newcite{riedel2010modeling} and \newcite{hoffmann2011knowledge} apply the multi-instance learning (MIL) mechanism for RE, which considers the reliability of each instance and combines multiple sentences containing the same entity pair together to alleviate the noise problem.

In recent years, neural models \cite{zhang2015relation,zeng2016incorporating,miwa2016end} have been widely used in RE. These neural models are capable of accurately capturing textual relations without explicit linguistic analysis. Based on these neural architectures and the MIL mechanism, \newcite{lin2016neural} propose a sentence-level attention to reduce the influence of incorrectly labeled sentences. To summarize, these MIL models generally make soft weight adjustment for informative and noisy instances. Some works further adopt external information to improve denoising performance: \newcite{ji2017distant} incorporate external entity descriptions to enhance attention representations; \newcite{liu2017soft} manually set label confidences to denoise entity-pair level noises.

More sophisticated mechanisms, such as reinforcement learning \cite{feng2018reinforcement,zeng2018large}, have recently also been adapted to select positive sentences from noisy data. However, these complex mechanisms usually require much time to fine-tune and the convergence is not yet well guaranteed in practice. In this paper, we propose a novel fine-grained denoising method for RE via adversarial training. The method is simple and effective to be applied in various neural architectures and to scale up to large-scale data.

\begin{figure*}[!htp]
\centering
\includegraphics[width=0.8\linewidth]{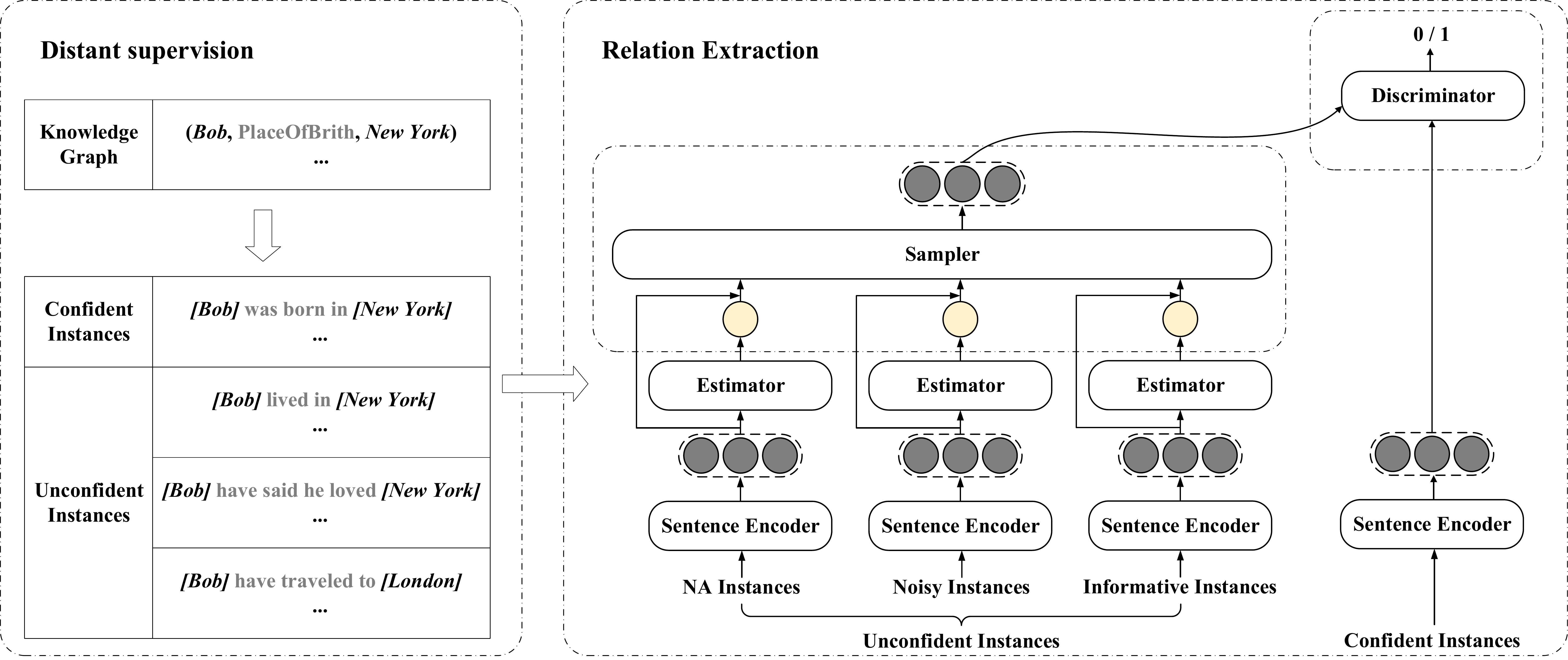}
\caption{The overall framework of the instance-level adversarial training model for relation extraction. The discriminator module is used to judge whether an instance is labeled correctly, and the instance will be considered coming from the confident set if the prediction is yes. The sampler module is used to select the most confusing instances from unconfident data to cheat the discriminator.}
\label{fig:framework}
\end{figure*}

\subsection{Adversarial Training}
\newcite{szegedy2013intriguing} propose to generate adversarial examples by adding noise in the form of small perturbations to the original data. These noise examples are often indistinguishable for humans but lead to models' wrong predictions. \newcite{goodfellow2014explaining} analyze adversarial examples and propose adversarial training for image classification tasks. Afterwards, \newcite{goodfellow2014generative} propose a mature adversarial training framework and use the framework to train generative models.

% \newcite{chen2017adversarial} and \newcite{liu2017adversarial} develop the adversarial training into multi-task learning for text classification and word segmentation. In text generation tasks, \newcite{yu2017seqgan} and \newcite{li2017adversarial} propose sequence generative adversarial nets for poem generation and dialogue generation respectively.

Adversarial training has also been explored in NLP. \newcite{miyato2015distributional} propose adversarial training for text classification by adding perturbations to word embeddings. The idea of perturbation addition has further been applied in other NLP tasks including language models \cite{xie2017data} and relation extraction \cite{wu2017adversarial}. Different from \cite{wu2017adversarial} that generates pseudo adversarial examples by adding perturbations to instance embeddings, we perform adversarial training by sampling adversarial examples from real-world noisy data. The adversarial examples in our method can better correspond to the real-world scenario for RE. Hence our method is more favorable to solve the wrong labeling problem in distant supervision, which will be shown in experiments.

\section{Methodology}

In this section, we introduce the details of our instance-level adversarial training model for denoising RE. For this model, we split the entire training data into two parts, the set of those confident instances $\mathcal{I}_c$ and the set of those unconfident instances $\mathcal{I}_{u}$. A sentence encoder is adopted to represent sentence semantics with embeddings. The adversarial training framework consists of a sampler and a discriminator, corresponding to the noise filter and the relation classifier respectively.

\subsection{The Framework}
F
As shown in Figure \ref{fig:framework}, the overall framework of our instance-level adversarial training model includes a discriminator $D$ and a sampler $S$, in which $S$ samples adversarial examples from the unconfident set $\mathcal{I}_{u}$, and $D$ learns to judge whether a given instance is from $\mathcal{I}_{c}$ or $\mathcal{I}_{u}$. 

We assume that each instance $s \in \mathcal{I}_c$ exposes implicit semantics of its labeled relation $r_s$. In contrast, those instances $s \in \mathcal{I}_{u}$ are not trusted to be labeled correctly during the adversarial training. Hence, we implement $D$ as a function $D(s, r_s)$ to judge whether a given instance $s$ exposes implicit semantics of its labeled relation $r_s$: if yes, the instance comes from $\mathcal{I}_c$; while if no, the instance comes from $\mathcal{I}_u$.

The training process is a min-max game and can be formalized as follows,
\begin{align}
\phi = \min_{p_{u}}\max_{D}(E_{s \sim p_c}[\log (D(s,r_s))] \\\nonumber
 + E_{s \sim p_{u}}[\log(1- D(s,r_s))]),
\end{align}
where $p_c$ is the confident data distribution, and the sampler $S$ samples adversarial examples from the unconfident data according to the probability distribution $p_u$. 

After sufficient training, $S$ tends to sample those informative instances in $\mathcal{I}_{u}$ rather than those noisy instances, and $D$ becomes a relation classifier of good robustness to noisy data. We will give the detailed introduction to the sampler in Section \ref{sec:sampler} and the discriminator in Section \ref{sec:discriminator}. 

\subsection{Sampler}
\label{sec:sampler}

The sampler module aims to select the most confusing sentences from the unconfident set $\mathcal{I}_{u}$ to cheat the discriminator as much as possible by optimizing the probability distribution $p_{u}$. Hence, we need to calculate the \emph{confusing score} for each instance in the unconfident set $\mathcal{I}_{u}$.

Given an instance $s$, we can use neural sentence encoders to represent its semantic information as an embedding $\mathbf{y}$. The details of neural encoders will be introduced in Section \ref{sec:encoder}. Here, we can simply calculate the confusing score according to the sentence embedding $\mathbf{y}$ as follows,
\begin{equation}
C(s) = \mathbf{W} \cdot \mathbf{y},
\end{equation}
where $\mathbf{W}$ is a separating hyperplane. We further define $P_{u}(s)$ as the \emph{confusing probability} over $\mathcal{I}_{u}$,
\begin{equation}
\label{eq:confusing}
P_{u}(s) = \frac{\exp(C(s))}{\sum_{s \in \mathcal{I}_{u}}\exp(C(s))}.
\end{equation}

In the unconfident set, we regard those instances with high $D(s, r_s)$ scores as the confusing instances, because they will fool the discriminator $D$ to make wrong decision. An optimized sampler will assign larger confusing score to those most confusing instances. Hence, we formalize the loss function to optimize the sampler module as follows:
\begin{equation}
\label{eq:op_sampler}
\mathcal{L}_S = - \sum_{s \in \mathcal{I}_{u}} P_{u}(s) \log(D(s, r_s)).
\end{equation}
When optimizing the sampler, we regard the component $P_{u}(s)$ as parameters for updating.

Note that, when an instance is labeled as $r_s = \texttt{NA}$, it indicates the relation of this instance is not available, either unsure or having no relation. Since these instances are always wrongly predicted into other relations, in order to let the discriminator restrain this tendency, we specifically define $D(s, \texttt{NA})$ as the average score of the instance over all feasible relations: 
\begin{equation}
D(s, \texttt{NA}) = \frac{1}{|\mathcal{R}|-1} \sum_{r \in \mathcal{R}, r \ne \texttt{NA}}D(s, r),
\end{equation}
where $\mathcal{R}$ indicates the set of relations.

\subsection{Discriminator}
\label{sec:discriminator}

Given an instance $s$ and its embedding $\mathbf{y}$, the discriminator is responsible for judging whether its labeled relation $r_s$ is correct. We implement the discriminator based on the semantic relatedness between $\mathbf{r}_s$ and $\mathbf{y}$,
\begin{equation}
D(s, r_s) = \sigma(\mathbf{r}_s \cdot \mathbf{y}),
\end{equation}
where $\sigma(\cdot)$ is the sigmoid function.

An optimized discriminator will assign high scores to those instances in $\mathcal{I}_c$ and low scores to those instances in $\mathcal{I}_u$. Hence, we formalize the loss function to optimize the discriminator module as follows:
\begin{align}
\label{eq:op_dis}
\mathcal{L}_D = -\sum_{s \in \mathcal{I}_c} \frac{1}{|\mathcal{I}_c|} \log (D(s, r_s)) \\\nonumber
 - \sum_{s \in \mathcal{I}_{u}} P_{u}(s) \log(1- D(s, r_s)).
\end{align}
When optimizing the discriminator, we regard the component $D(s, r_s)$ as parameters for updating. Note that, the objective functions of the sampler in Eq. \ref{eq:op_sampler} and the discriminator in Eq. \ref{eq:op_dis} are adversarial to each other.

In practice, the data set is usually too large to be frequently traversed due to intractable large amounts of computation. For convenience of training efficiency, we can simply sample subsets to approximate the probability distribution. Hence, we formalize a new loss function for optimization:
\begin{align}
\tilde{\mathcal{L}}_D = -\sum_{s \in \hat{\mathcal{I}}_c} \frac{1}{|\hat{\mathcal{I}}_c|} \log (D(s, r_s)) \\\nonumber
 - \sum_{s \in \hat{\mathcal{I}}_{u}} Q_{u}(s) \log(1- D(s,r_s)),
\end{align}
where $\hat{\mathcal{I}}_c$ and $\hat{\mathcal{I}}_{u}$ are subsets sampled from $\mathcal{I}_c$ and $\mathcal{I}_{u}$ respectively, and $Q_{u}(s)$ is the corresponding approximation to $P_u(s)$ in Eq. 
\ref{eq:confusing}: 
\begin{equation}
Q_{u}(s) = \frac{\exp(C(s)^{\alpha})}{\sum_{s \in \hat{\mathcal{I}}_{u}}\exp(C(s)^{\alpha})}.
\end{equation}
Note that $\alpha$ is a hyper-parameter that controls the sharpness of the confusing probability distribution. For consistency, we also approximate $\mathcal{L}_S$ in Eq. \ref{eq:op_sampler} as:
\begin{align}
\tilde{\mathcal{L}}_S = - \sum_{s \in \hat{\mathcal{I}}_{u}} Q_{u}(s) \log(D(s, r_s)).
\end{align}
$\tilde{\mathcal{L}}_S$ and $\tilde{\mathcal{L}}_D$ are used to optimize our adversarial training model.

\subsection{Instance Encoder}
\label{sec:encoder}
Given an instance $s$ containing two entities, we apply several neural network architectures to encode the sentence into continuous low-dimensional embeddings $\mathbf{y}$, which are expected to capture the implicit semantics of the labeled relation between two entities.

\subsubsection{Input Layer}
The input layer aims to map discrete language symbols (i.e., words) into continuous input embeddings. Given an instance $s$ containing $n$ words $\{w_1, \ldots, w_n\}$, we use Skip-Gram \cite{mikolov2013efficient} to embed all words into $k_w$-dimensional space $\{\mathbf{w}_1, \ldots, \mathbf{w}_n\}$. For each word $w_i$, we also embed its relative distances to the two entities into two $k_p$-dimensional vectors, and then concatenate them as an unified position embedding $\mathbf{p}_i$ \cite{zeng2014relation}. We finally get the $k_i$-dimensional input embeddings for the following encoding layer,
\begin{align}
\mathbf{s} &= \{\mathbf{x}_1,\ldots, \mathbf{x}_n\} \\\nonumber
&= \{[\mathbf{w}_1;\mathbf{p}_1],\ldots, [\mathbf{w}_n;\mathbf{p}_n]\}.
\end{align}

\subsubsection{Encoding Layer}
In the encoding layer, we select four typical architectures including CNN \cite{zeng2014relation}, PCNN \cite{zeng2015distant}, RNN \cite{zhang2015relation} and BiRNN \cite{zhang2015relation} to further encode input embeddings of the instance into sentence embeddings. 

\textbf{CNN} slides a convolution kernel with the window size $m$ over the input sequence $\{\mathbf{x}_1,\ldots, \mathbf{x}_n\}$ to get the $k_h$-dimensional hidden embeddings. 
\begin{align}
\mathbf{h}_i &= \text{CNN}\big( \mathbf{x}_{i - \frac{m-1}{2}}, \ldots ,\mathbf{x}_{i + \frac{m-1}{2}} \big).
\end{align}
A max-pooling is then applied over these hidden embeddings to output the final instance embedding $\mathbf{y}$ as follows,
\begin{align}
[\mathbf{y}]_{j} &= \max \{[\mathbf{h}_{1}]_{j}, \ldots, [\mathbf{h}_{n}]_{j} \}.
\end{align}

\textbf{PCNN} is an extension to CNN, which also adopts a convolution kernel with the window size $m$ to obtain hidden embeddings. Afterwards, PCNN divides the hidden embeddings into three segments $\{\mathbf{h}_1,\ldots,\mathbf{h}_{e_1}\}$, $\{\mathbf{h}_{e_1+1},\ldots,\mathbf{h}_{e_2}\}$, and $\{\mathbf{h}_{e_2+1},\ldots,\mathbf{h}_{n}\}$, where $e_1$ and $e_2$ are entity positions. PCNN applies a piecewise max-pooling for each segment,
\begin{align}
[\mathbf{y}_1]_{j} &= \max \{[\mathbf{h}_{1}]_{j}, \ldots, [\mathbf{h}_{e_1}]_{j} \},\\\nonumber
[\mathbf{y}_2]_{j} &= \max \{[\mathbf{h}_{e_1+1}]_{j}, \ldots, [\mathbf{h}_{e_2}]_{j} \},\\\nonumber
[\mathbf{y}_3]_{j} &= \max \{[\mathbf{h}_{e_2+1}]_{j}, \ldots, [\mathbf{h}_{n}]_{j} \}.
\end{align}
By concatenating all pooling results, PCNN eventually outputs a $3 \cdot k_h$-dimensional instance embedding $\mathbf{y}$ as follows,
\begin{align}
\mathbf{y} = [\mathbf{y}_1; \mathbf{y}_2; \mathbf{y}_3].
\end{align}

\textbf{RNN} is designed for modeling sequential data, as it keeps its hidden state changing with input embeddings at each time-step accordingly,
\begin{align}
\mathbf{h}_i = \text{RNN}(\mathbf{x}_i, \mathbf{h}_{i-1}),
\end{align}
where $\text{RNN}(\cdot)$ is the recurrent unit and $\mathbf{h}_i \in \mathbb{R}^{k_h}$ is the hidden embedding at the time-step $i$. In this paper, we select gated recurrent unit (GRU) \cite{cho2014properties} as the recurrent unit. We use the hidden embedding of the last time-step as the instance embedding, i.e., $\mathbf{y} = \mathbf{h}_n$.

\textbf{Bi-RNN} aims to incorporate information from both sides of the sentence sequence. Bi-RNN is adopted with forward and backward directions as follows,
\begin{align}
\overrightarrow{\mathbf{h}}_i = \text{RNN}_{f}(\mathbf{x}_i, \overrightarrow{\mathbf{h}}_{i-1}),\\\nonumber
\overleftarrow{\mathbf{h}}_i = \text{RNN}_{b}(\mathbf{x}_i, \overleftarrow{\mathbf{h}}_{i+1}),
\end{align}
where $\overrightarrow{\mathbf{h}}_i$ and $\overleftarrow{\mathbf{h}}_i$ are the hidden states at the position $i$ of the forward and backward RNN respectively. We concatenate the hidden states from both the forward and backward RNN as the instance embedding $\mathbf{y}$,
\begin{align}
\mathbf{y} = [\overrightarrow{\mathbf{h}}_n; \overleftarrow{\mathbf{h}}_1].
\end{align}

\subsection{Initialization and Implementation Details}
Here we introduce the learning and optimization details for our adversarial training model. We define the optimization function as 
\begin{equation}
\mathcal{L} =  \tilde{\mathcal{L}}_D + \lambda \tilde{\mathcal{L}}_S,
\end{equation}
where $\lambda$ is a harmonic factor. In practice, both the modules in adversarial training are optimized alternately using stochastic gradient descent (SGD). Since the framework of our model is much simpler than typical generative adversarial networks (GAN), we do not have to calibrate alternating ratio between the loss functions, and hence we can simply use a $1:1$ ratio. It enables our model efficient for learning on large-scale data. Moreover, we can also integrate $\lambda$ into the learning rate of the sampler $\tilde{\mathcal{L}}_S$, so as to avoid adjusting the hyper-parameter $\lambda$.

At the start of adversarial training, we pre-train a relation classifier on the entire training data. The relation classifier will split the entire data into a small confident data and a large unconfident data. During the adversarial training, after every few training epochs, some instances from the unconfident set that are both recommended by the sampler and recognized by the discriminator will be selected to enrich the confident set. 

\begin{figure*}[t]
\centering
\subfigure[Comparison of the proposed models and feature-based models.]{
\label{fig:Fig1}
\includegraphics[width=0.32\textwidth]{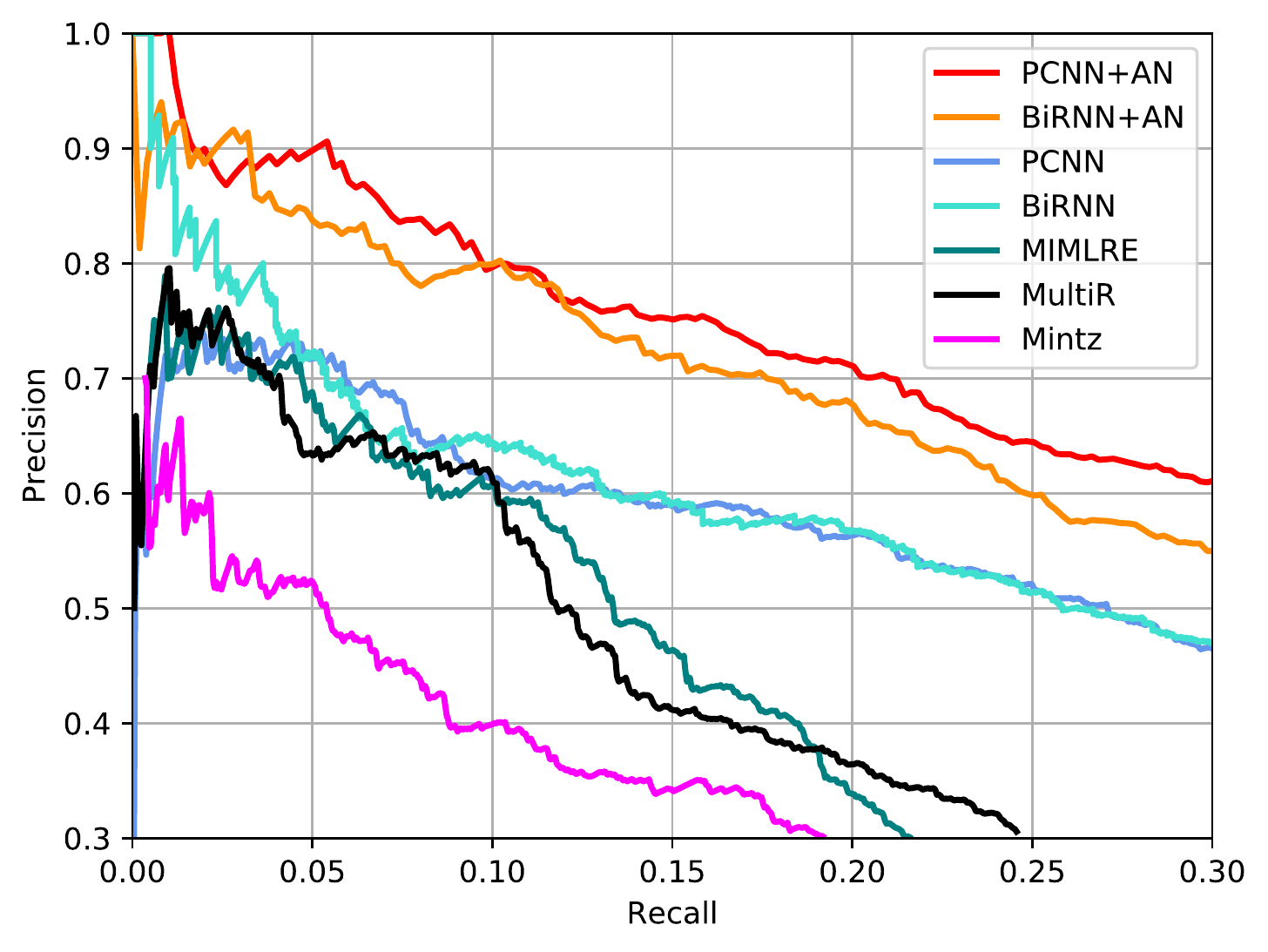}}
\subfigure[Comparison of the proposed models and various CNN models.]{
\label{fig:Fig2}
\includegraphics[width=0.32\textwidth]{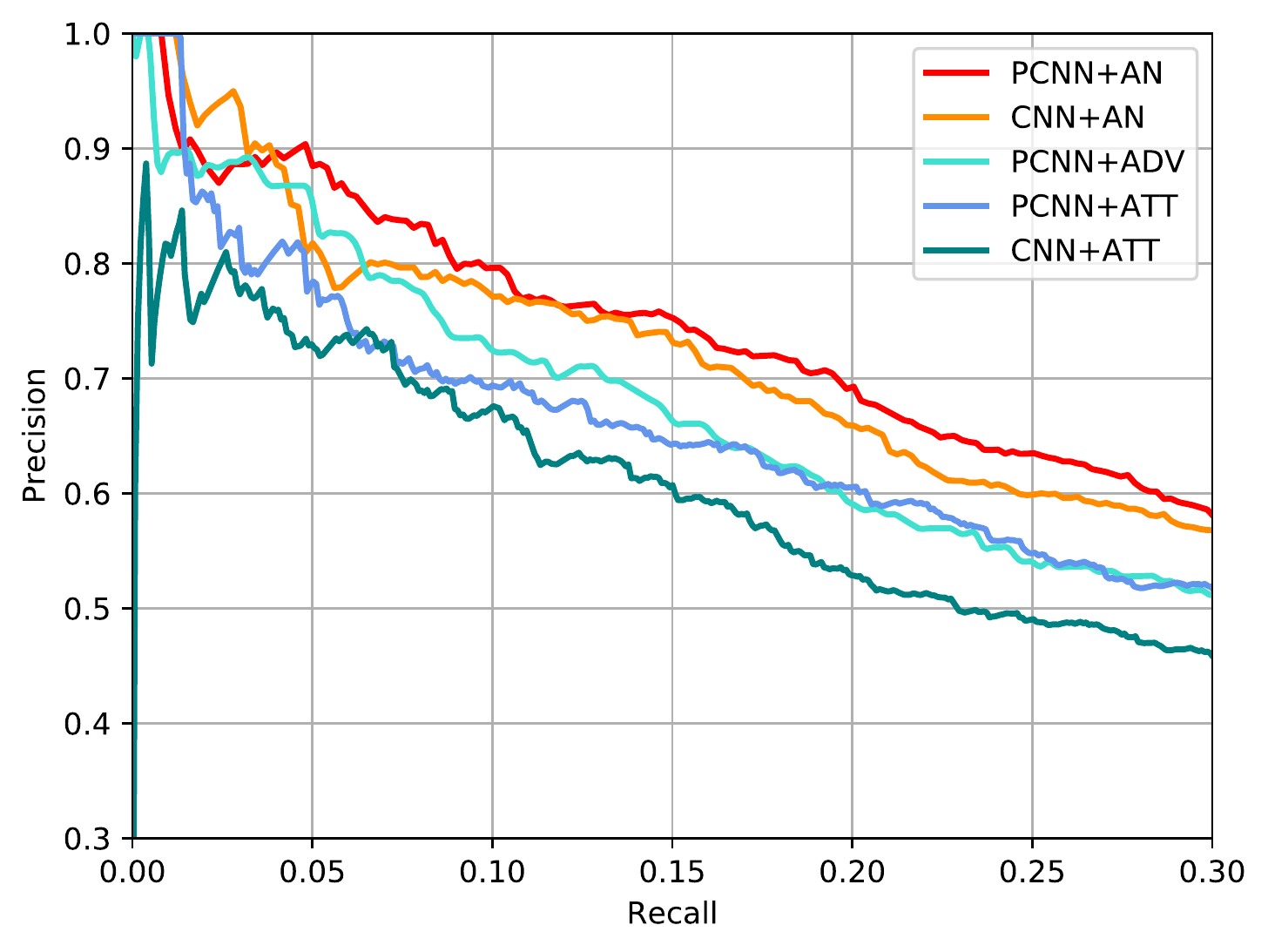}}
\subfigure[Comparison of the proposed models and various RNN models.]{
\label{fig:Fig3}
\includegraphics[width=0.32\textwidth]{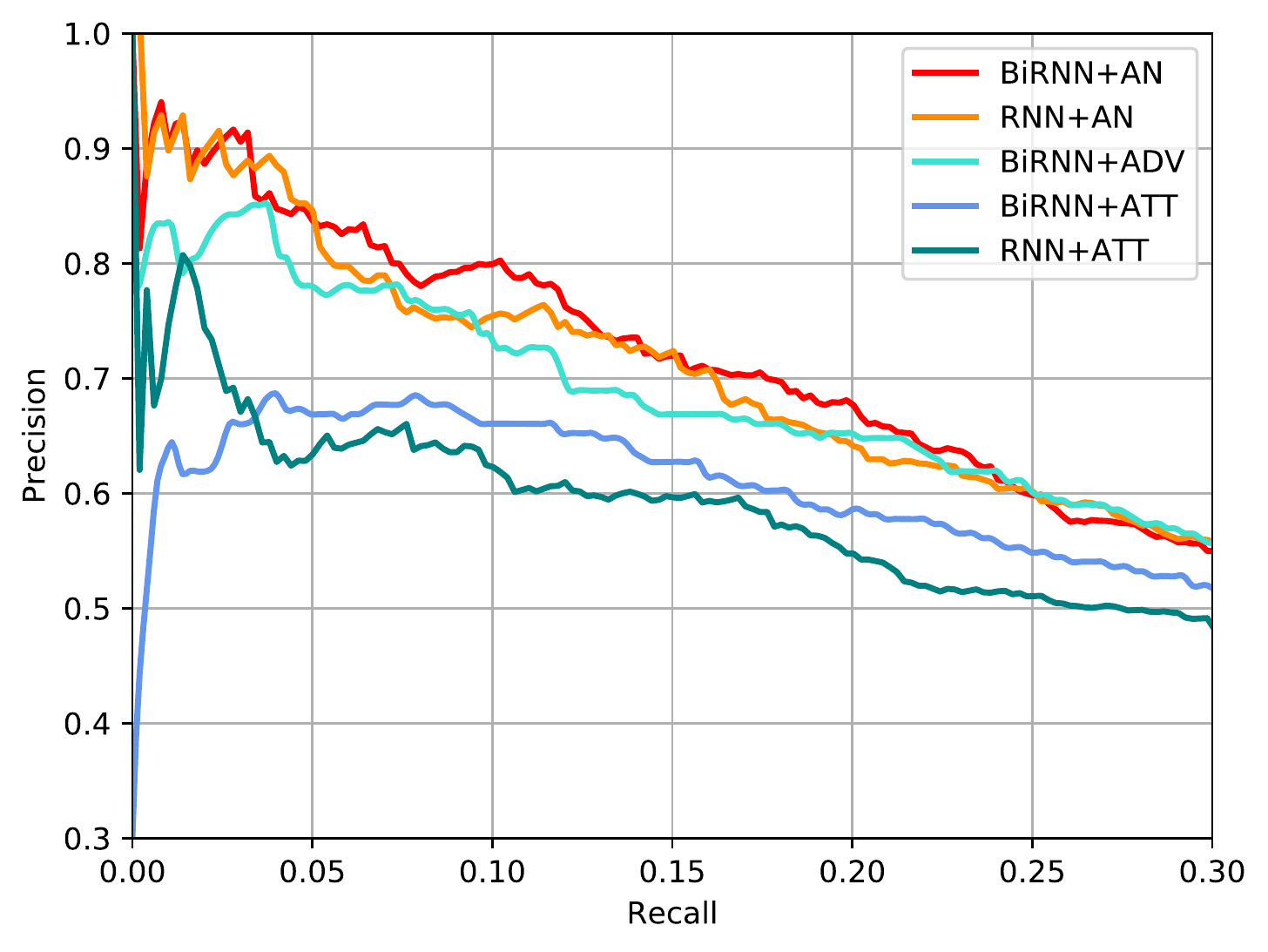}}
\caption{Aggregate precision/recall curves of different models.}
\label{fig:overallresults}
\end{figure*}

\section{Experiments}
% Our experiments are intended to demonstrate that our adversarial denoising method can conduct fine-grained denoising operations and further alleviate the wrong labeling problem for RE models. 

In this section, we carry out experiments to demonstrate the effectiveness of our instance-level adversarial training method. We first introduce datasets and parameter settings. Afterwards, we compare the performance of our method with conventional neural methods and feature-based methods for RE. To further verify that our method can better discriminate those informative instances from noisy ones, we also conduct evaluations on those entity pairs with few sentences.

\subsection{Datasets and Experiment Settings}

\subsubsection{Datasets}

We conduct experiments on the benchmark dataset derived from New York Times (NYT) corpus, which is first proposed by \newcite{mintz2009distant} and then widely used in various distantly supervised RE works \cite{riedel2010modeling,hoffmann2011knowledge,surdeanu2012multi,zeng2015distant,lin2016neural,wu2017adversarial}. The dataset aligns entity pairs and their relations in the KG Freebase with NYT corpus. After various essential data processing, there are $53$ relation types including the \texttt{NA} relation in this dataset. The training data contains $522,611$ sentences, $281,270$ entity pairs and $18,252$ relational facts. The test data contains $172,448$ sentences, $96,678$ entity pairs and $1,950$ relational facts.

\subsubsection{Parameter Settings}
In our models, we select the learning rate $\alpha_d$ and $\alpha_s$ among $\{0.5, 0.1, 0.05, 0.01\}$ for training the discriminator and the sampler respectively. For other parameters, we simply follow the settings used in \cite{zeng2014relation,lin2016neural,wu2017adversarial} so that we can fairly compare the results of our adversarial denoising models with these baselines. Table \ref{parameters} shows all parameters used in the experiments. During training, we select most informative and confident instances in the unconfident set to enrich the confident set every $10$ training epochs.

\begin{table}[!htp]
\centering
\scalebox{0.9}{
\begin{tabular}{c|c}
\toprule
\multicolumn{1}{l|}{Discriminator Learning Rate $\alpha_d$}        & 0.1 \\
\multicolumn{1}{l|}{Sampler Learning Rate $\alpha_s$}        & 0.01 \\
\multicolumn{1}{l|}{Hidden Layer Dimension $k_h$ for CNNs}        & 230   \\
\multicolumn{1}{l|}{Hidden Layer Dimension $k_h$ for RNNs}        & 150   \\
\multicolumn{1}{l|}{Position Dimension $k_p$ for CNNs}   & 5     \\
\multicolumn{1}{l|}{Position Dimension $k_p$ for RNNs}   & 3     \\
\multicolumn{1}{l|}{Word Dimension $k_w$} & 50    \\
\multicolumn{1}{l|}{Convolution Kernel Size $m$}    & 3     \\
\multicolumn{1}{l|}{Dropout Probability $p$}            & 0.5  \\
\bottomrule
\end{tabular}
}
\caption{Parameter settings.}
\label{parameters}
\end{table}

\subsection{Overall Evaluation Results}

We follow \newcite{mintz2009distant} to conduct the held-out evaluation. We construct candidate triples by combining entity pairs in the test set with various relations and rank these triples according to their corresponding sentence representations. By regarding the triples in the KGs as correct and others as incorrect, we evaluate different models with their precision-recall results.

The evaluation results are shown in Figure \ref{fig:overallresults} and Table \ref{tab:overallresults}. We report the results of various neural architectures including CNN, PCNN, RNN and BiRNN with various denoising methods: \textbf{+ATT} is the selective attention method over instances \cite{lin2016neural}; \textbf{+ADV} is the denoising method by adding a small adversarial perturbation to instance embeddings \cite{wu2017adversarial}; \textbf{+AN} is our proposed adversarial training method. We also compare our methods with feature-based models, including Mintz \cite{mintz2009distant}, MultiR \cite{hoffmann2011knowledge} and MIML \cite{surdeanu2012multi}. The results of the baseline models all come from the data reported in their papers or their open-source code. From the figure and table, we observe that: 

(1) As shown in Figure \ref{fig:Fig1}, neural models significantly outperform all feature-based models over the entire range of recall. When the recall gradually grows, the performance of feature-based models drops out quickly. However, all the neural models still preserve stable and competitive precision. It demonstrates that human-designed features cannot work well in a noisy environment, and inevitable errors brought by NLP tools will further hurt the performance. In contrast, instance embeddings learned automatically by neural models can effectively capture implicit relational semantics from noisy data for RE.

\begin{table}[t]
\centering
\scalebox{0.9}{
\begin{tabular}{ll|c|c|c|c} 
\toprule
\multicolumn{2}{c|}{Method} & 0.1 & 0.2 & 0.3 & Mean \\
\midrule
\multirow{2}{*}{CNN+}				&ATT& 67.5 	& 52.8 & 45.8&55.4\\
									&AN& \textbf{75.3}& \textbf{66.3} &\textbf{54.3}&\textbf{65.3}\\
\midrule									
\multirow{2}{*}{RNN+}				&ATT&63.9 & 54.4 & 48.0 & 55.4\\
									&AN&\textbf{75.3} & \textbf{64.5} & \textbf{55.8} & \textbf{65.2}\\
\midrule									
\multirow{3}{*}{PCNN+}				&ATT&69.4 & 60.6 & 51.6 & 60.5 \\
									&ADV&71.7 & 58.9 & 51.1 & 60.6 \\ 
									&AN&\textbf{80.3} & \textbf{70.2} & \textbf{60.3} & \textbf{70.3} \\
\midrule									
\multirow{3}{*}{BiRNN+}				&ATT&66.8 & 58.6 & 52.4 & 64.2 \\
									&ADV&72.8 & 64.6 & \textbf{55.3} & 65.2 \\
									&AN&\textbf{79.1} & \textbf{67.3} & 54.1 & \textbf{66.8} \\
\bottomrule
\end{tabular}}
\caption{Precision of various models for different recall (\%).}
\label{tab:overallresults}
\end{table}

(2) Both for CNNs (CNN and PCNN) in Figure \ref{fig:Fig2} and RNNs (RNN and BiRNN) in Figure \ref{fig:Fig3}, the models with adversarial training outperform the models with sentence-level attention. The sentence-level attention over multiple instances, which calculates soft weights for each sentence to reduce noise, only makes a coarse-grained distinction between informative and noisy instances. In contrast, the neural models trained with adversarial denoising methods generate or sample noisy adversarial examples and force the relation classifiers to overcome them. Hence, the models with adversarial training provide efficient noise reduction in finer granularity. In general, the models with our adversarial training method achieve the best results among models using adversarial training. This indicates that, as compared to generating pseudo adversarial examples by adding perturbations, our method by sampling adversarial examples from real-world instances can better discriminate informative instances from noisy instances.

(3) To better compare various denoising methods, we also show evaluation results in Table \ref{tab:overallresults}. Since we focus more on the performance of those top-ranked results, here we show the precision scores when the recall is $0.1$, $0.2$, $0.3$ as well as their mean. We find that complicated neural models (PCNN, BiRNN) perform better than simple neural networks (CNN, RNN) when using the same denoising methods. Both CNNs and RNNs are significantly improved by adversarial training, and our method (AN) performs consistently much better than the adversarial training baseline (ADV). The improvements brought by changing denoising methods are more significant than the improvements brought by modifying neural models. This indicates that the wrong labeling problem is the critical factor that prevents distantly supervised RE models from working effectively.

\begin{table*}[t]
\centering
\scalebox{0.9}{
\begin{tabular}{lcccccccccccc}
\toprule
\multicolumn{1}{l}{{Test Settings}} & \multicolumn{4}{c}{{One}}                                       & \multicolumn{4}{c}{{Two}}                                       & \multicolumn{4}{c}{{All}}                                       \\
\cmidrule(lr){2-5} \cmidrule(lr){6-9} \cmidrule(lr){10-13}
\multicolumn{1}{l}{P@N}&\multicolumn{1}{c}{100}&\multicolumn{1}{c}{200}&\multicolumn{1}{c}{300}&\multicolumn{1}{c}{Mean}&\multicolumn{1}{c}{100}&\multicolumn{1}{c}{200}&\multicolumn{1}{c}{300}&\multicolumn{1}{c}{Mean}&\multicolumn{1}{c}{100}&\multicolumn{1}{c}{200}&\multicolumn{1}{c}{300}&\multicolumn{1}{c}{Mean}\\
\midrule
PCNN & 63.0 & 61.0 & 55.3 & 59.8 &65.0 &62.5 &57.3 & 61.6 & 71.0 &64.0 &58.7 & 64.6 \\
PCNN+ONE      & 73.3          & 64.8          & 56.8          & 65.0          & 70.3          & 67.2          & 63.1          & 66.9          & 72.3          & 69.7          & 64.1          & 68.7          \\
PCNN+AVG      & 71.3          & 63.7          & 57.8          & 64.3          & 73.3          & 65.2          & 62.1          & 66.9          & 73.3          & 66.7          & 62.8          & 67.6          \\
PCNN+ATT      & 73.3          & 69.2          & 60.8          & 67.8          & 77.2          & 71.6          & 66.1          & 71.6          & 76.2          & 73.1          & 67.4          & 72.2          \\
%PCNN+SL       & 84.0          & \textbf{75.5} & 68.3          & 75.9          & 86.0          & 77.0          & 73.3          & 78.8          & 87.0          & \textbf{84.5} & 77.0          & 82.8          \\
%PCNN+D        & -             & -             & -             & -             & -             & -             & -             & -             & 87.0          & 83.0          & 74.0          & 81.3          \\
\midrule
PCNN+AN       & \textbf{84.0} & \textbf{75.0} & \textbf{73.0} & \textbf{77.3} & \textbf{86.0} & \textbf{77.0} & \textbf{73.7} & \textbf{78.9} & \textbf{90.0} & \textbf{82.0} & \textbf{76.3} & \textbf{82.8}\\
\bottomrule
\end{tabular}}
\caption{Top-N precision (P@N) for RE in the entity pairs with different number of instances (\%).}
\label{tab:pnresults}
\end{table*}

\subsection{Effect of Adversarial Denoising Training}
% Recently, there are also some other MIL denoising methods proposed based on the sentence-level attention model.

To further verify the effectiveness of our adversarial training method, we evaluate the RE performance of our method and conventional MIL denoising methods in a more challenging scenario, i.e., when entity pairs having few sentences. 

For each entity pair, we randomly select one sentence, two sentences, and all sentences to construct three experimental settings respectively. We report P@100, P@200, P@300 and the mean of them in the held-out evaluation. Since PCNN is the best neural model in the above comparison, we simply use PCNN to compare our method (AN) with the recent state-of-the-art denoising method, sentence-level attention (ATT), as well as its naive versions \textbf{+ONE} and \textbf{+AVG} \cite{zeng2015distant,lin2016neural}. The evaluation results are shown in Table \ref{tab:pnresults}, and from the results we observe that: 

(1) Our method achieves consistent and significant improvements as compared to the ATT method and its naive versions, especially when each entity pair only corresponds to one or two sentences. The reason is that most MIL denoising methods including ATT typically assume that at least one instance that mentions the given entity pair can express their relation, and always select at least one informative sentence for the entity pair. This assumption is not always true especially when entity pairs correspond to few sentences: it is more likely there is no instance that can express the relation of the given entity pair. In contrast, our adversarial training method is not restricted by the assumption. By conducting on instance level individually, our method keeps effective even when the instances of each entity pair are few.

(2) When taking more instances into account, all models achieve better results. PCNN+ATT and PCNN+AN achieve more improvements than those naive methods. The growth of distant supervision data brings more information for training RE models as well as more noises that may hurt performance. Our method keeps its degree of superiority to the ATT method as the data growth. This indicates that our method could provide more robust and reliable scheme to denoise distant supervision data.

% PCNN+SL incorporates an additional confidence evaluation system guided by manual thresholds to obtain soft labels for sentences. PCNN+D uses a complicated mechanism to embed extra entity description information for noise reduction. These methods can make use of extra outside information when there are only insufficient sentences and indeed work very well. However, the extra information such as entities' descriptions is not available at any time. As compared with these models, our model still has the comparable results. And the fewer the sentences are, the better results our model can achieve. Without incorporating extra outside information, our model also achieves great results. This means that our denoising method better distinguishes noisy and informative components from distantly supervised datasets. The performance of our model can be further improved if we incorporate extra information.

\begin{table}[t]
\centering \small
\scalebox{0.9}{
\begin{tabular}{c|p{0.8\columnwidth}}
\toprule
{Relation} & \texttt{Location Contains} \\
\midrule
\multirow{2}{*}{{Positive}} & ... China's 10 most polluted cities, four, including \textbf{Datong}, are in \textbf{Shanxi} province ... \\
\cmidrule(lr){2-2}
& ... \textbf{Manhattan}'s \textbf{Chinatown} has fought off the forces of urban decline ... \\
\midrule
\midrule
\multirow{2}{*}{{Negative}}  & ... the senior commander of U.S. forces in \textbf{Baghdad}, has figured out the obstacle to america 's dream for \textbf{Iraq} ...\\
\cmidrule(lr){2-2}
& ... after \textbf{Japan}'s defeat, he said, American soldiers drove jeeps onto his family 's estate in \textbf{Iwate} ...\\
\bottomrule
\end{tabular}}
\caption{Some examples sampled by the sampler in NYT corpus.}
\label{tab:casestudy}
\end{table}

\subsection{Case Study}

Table \ref{tab:casestudy} shows examples sampled by the sampler. For the frequent relation \texttt{Location Contains}, we use the sampler to select the positive and negative instances respectively. For each sentence, we highlight the entities in boldface. From the table we find that: The former positive examples clearly correspond to the relation \texttt{Location Contains}, while those negative examples fail to reflect this relation. These examples show that our sampler is effective to distinguish informative and noisy instances.

\section{Conclusion}

In this paper, we propose a denoising distant supervised method for RE via instance-level adversarial training. By splitting the entire data into the confident set and the unconfident set, our method trains a sampler and a discriminator adversarially. The sampler aims to select the most confusing instance from the unconfident set, and the discriminator aims to distinguish an instance which comes from either the confident set or the unconfident set. In experiments, we apply our method to various neural architectures for RE. The experimental results show that our method achieves efficient noise reduction in finer granularity and significantly outperforms the state-of-the-art baseline. Our method is also robust for those long-tail entity pairs with few instances.

In the future, we plan to explore the following directions: (1) Inspired by \cite{ji2017distant}, it will be promising to adopt external knowledge, from either KBs or text, to help train more efficient samplers and discriminators for adversarial training. (2) We may also extend the instance-level adversarial training to the entity-pair level to further improve the robustness of RE models.

\bibliography{acl2018}

\end{document}